\documentclass[sigconf]{acmart}
\copyrightyear{2026}
\acmYear{2026}
\setcopyright{cc}
\setcctype{by}
\acmConference[MM '26]{Proceedings of the 34th ACM International Conference on Multimedia}{November 10--14, 2026}{Rio de Janeiro, Brazil}
\acmBooktitle{Proceedings of the 34th ACM International Conference on Multimedia (MM '26), November 10--14, 2026, Rio de Janeiro, Brazil}
\acmDOI{10.1145/3767308.3835199}
\acmISBN{979-8-4007-2213-4/2026/11}

\begin{document}

\title{Slot-ID: Identity-Preserving Video Generation from Reference Videos via Slot-Based Temporal Identity Encoding}

\author{Yixuan Lai}
\orcid{0009-0006-3308-2745}
\email{yixuan.lai@zju.edu.cn}
\affiliation{%
  \institution{State Key Lab of CAD\&CG}
  \institution{Zhejiang University}
  \city{Hangzhou}
  \country{China}
}

\author{He Wang}
\orcid{0000-0002-2281-5679}
\email{realcrane@gmail.com}
\affiliation{%
  \institution{Department of Computer Science}
  \institution{University College London}
  \city{London}
  \country{UK}
}

\author{Kun Zhou}
\orcid{0000-0003-4243-6112}
\email{kunzhou@acm.org}
\affiliation{%
  \institution{State Key Lab of CAD\&CG}
  \institution{Zhejiang University}
  \city{Hangzhou}
  \country{China}
}

\author{Tianjia Shao}
\authornote{Corresponding author.}
\orcid{0000-0001-5485-3752}
\email{tjshao@zju.edu.cn}
\affiliation{%
  \institution{State Key Lab of CAD\&CG}
  \institution{Zhejiang University}
  \city{Hangzhou}
  \country{China}
}

\renewcommand{\shortauthors}{Yixuan Lai, He Wang, Kun Zhou, \& Tianjia Shao}

\begin{abstract}
Human identity-preserving text-to-video generation remains challenging under large changes in viewpoint, facial expression, illumination, and motion. Existing methods condition the generator on a single reference portrait, but a static image cannot capture how identity-bearing cues evolve across views and expressions, leading to face deformation, pose locking, identity drift, or over-smoothed faces.
We observe that a short reference clip naturally provides richer temporal and multi-view identity cues than any single image, motivating a video-referential formulation. This richer signal, however, introduces a new challenge: identity evidence is distributed across many frames and must be distilled into a compact, stable representation under a limited token budget.
To this end, we propose \textbf{Slot-ID}, a lightweight identity-conditioning framework built on a frozen text-to-video backbone. Slot-ID employs a slot-based temporal identity encoder with Sinkhorn-routed iterative reading to distill a compact, stable set of identity tokens from the reference clip, complemented by an image-anchor stream for dual-source conditioning.
Extensive experiments demonstrate that Slot-ID outperforms state-of-the-art methods in identity preservation and visual naturalness while remaining competitive in prompt following, with particularly large gains under challenging pose, expression, and motion variations. The project page is available at \url{https://gapszju.github.io/SlotID}.
\end{abstract}



\begin{teaserfigure}
  \includegraphics[width=\textwidth]{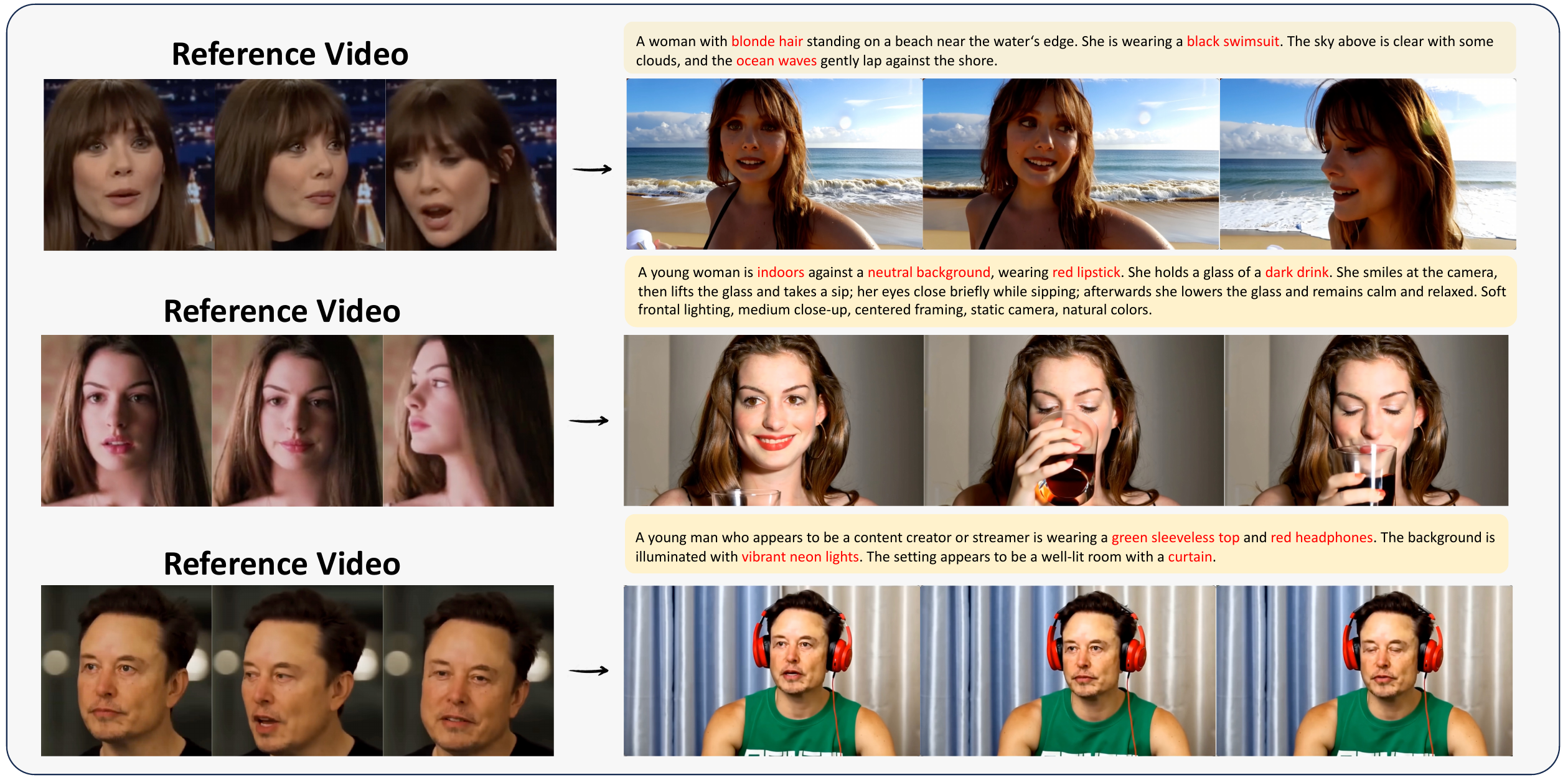}
  \caption{\textbf{Identity-preserving text-to-video generation with Slot-ID.}
  From a reference video (left), we compute a slot-based temporal identity code to condition generation. The synthesized clips (right) follow diverse instructions---changing actions, scenes, and apparel---while consistently preserving the subject's identity; colored text highlights key attributes.}
  \Description{The teaser shows a reference video on the left and several generated video examples on the right. The generated examples depict the same subject under different prompts, actions, scenes, and clothing while maintaining consistent facial identity.}
  \label{fig:teaser}
\end{teaserfigure}


\maketitle

\section{Introduction}
\label{sec:intro}
Generating realistic human videos from text prompts (T2V) has attracted growing research attention in recent years~\cite{sora,yang2024cogvideox,kong2024hunyuanvideo,wan2025wan,he2024idanimatorzeroshotidentitypreservinghuman,wei2025echovideoidentitypreservinghumanvideo,yuan2025identitypreservingtexttovideogenerationfrequency,xue2025standinlightweightplugandplayidentity,liu2025phantomsubjectconsistentvideogeneration,hu2025hunyuancustommultimodaldrivenarchitecturecustomized,huang2023vbenchcomprehensivebenchmarksuite,lin2024open}, opening up applications in content creation, virtual humans, and personalized media generation~\cite{wei2023dreamvideocomposingdreamvideos,jiang2024videobooth,materzynska2024newmove,li2024photomaker,xue2025standinlightweightplugandplayidentity,liu2025phantomsubjectconsistentvideogeneration,10.1145/3613904.3642575,IAB-SoD-2025,hollister-2024-toysrus-sora,kodaira2025streamditrealtimestreamingtexttovideo}. Yet preserving a user-specified human identity throughout a generated video remains difficult. Unlike personalized image generation, identity-preserving T2V requires the model to maintain an identifiable face across changes in views, expressions, illumination and motion, while still producing temporally coherent and prompt-faithful video. Achieving this balance remains an open challenge.

A common practice in current identity-preserving video generation is to extract identity cues from a single reference portrait and inject them into a pretrained backbone. This line of work is attractive for its simplicity, but it is fundamentally limited by the information available in a single image. In many single-image pipelines, identity is represented by features extracted from ArcFace-, CLIP-, or image-encoder-based models~\cite{radford2021learningtransferablevisualmodelsCLIP,Deng_2022Arcface,he2024idanimatorzeroshotidentitypreservinghuman,wei2025echovideoidentitypreservinghumanvideo,yuan2025identitypreservingtexttovideogenerationfrequency}. Such features are effective for identity matching, but they often underrepresent the temporally varying and view-dependent cues required for stable video synthesis. Image-only conditioning also tends to collapse under large pose or expression changes. Hence these methods often yield apparent artifacts such as face deformation and identity drift (Fig. \ref{fig:qualitative}), as well as pose locking (see supplementary material). As a result, these methods struggle precisely in the scenarios that matter most for video: large viewpoint variation, expressive motion, and sustained temporal consistency.

We argue that the missing ingredient is not simply stronger image features, but a richer reference signal. A single portrait cannot reveal how identity-bearing cues evolve across time, viewpoint, and expression. By contrast, a short reference video contains subject-specific temporal evidence: how the face moves when smiling, how appearance changes under mild viewpoint shifts, and which visual traits remain stable across frames despite motion and illumination variation. These cues are often crucial for distinguishing a person beyond a static frontal appearance. This suggests that identity-preserving T2V should be formulated not only as conditioning on single appearance, but as conditioning on a compact representation of temporally distributed identity evidence.

However, introducing a reference clip does not automatically solve the problem. Video conditioning encounters a new challenge: identity evidence is now distributed across many frames, mixed with redundancy, background context, and frame-specific variation, yet it must still be distilled into a small set of conditioning tokens compatible with a frozen video backbone. Naively encoding all clip features is computationally expensive and, in our experiments, tends to produce unstable or over-smoothed identities under strong viewpoint or expression changes. The key question is therefore how to aggregate person-specific evidence from a short clip into a representation that is both compact and stable enough to guide the generation effectively.

% \begin{figure}[t]
%   \centering
%   \includegraphics[width=\linewidth]{figs/intro_figs/issue.pdf}
%   \caption{\textbf{Failures from single-image references.}
%   \textbf{Left:} reference portraits.
%   \textbf{Top-right:} \emph{Face deformation}: viewpoint changes warp facial geometry (e.g., stretched cheeks/jawline and misaligned eyes).
%   \textbf{Bottom-right:} \emph{Pose locking}: a near-frontal yaw/pitch persists across frames, resulting in puppet-like motion.}
%   \label{fig:issue}
% \end{figure}

To address this, we propose \textbf{Slot-ID}, an identity-conditioning framework for text-to-video generation from short reference videos. 
We formulate and instantiate a video-referential identity conditioning problem: identity evidence distributed across a reference clip must be compressed into a compact, stable, backbone-compatible condition for T2V generation.
Our central idea is therefore to encode the reference clip into a small set of temporal identity tokens that summarize complementary identity cues across frames, rather than relying on a single pooled descriptor or a static portrait embedding alone. Concretely, we employ a slot-based temporal identity encoder, where learnable slots selectively aggregate identity-relevant evidence from the spatiotemporal reference features under a fixed token budget. 
To make this aggregation stable and non-redundant, we use Sinkhorn-routed assignment together with iterative refinement, which prevents a few locally strong responses from dominating all slots and improves consistency across neighboring frames. In parallel, we retain a clean image anchor---an almost frontal portrait frame from the video---to complement the video stream, combining stable appearance cues with temporal and multi-view identity evidence. The resulting identity condition can be injected into a frozen T2V backbone with minimal modification while remaining compatible with its pretrained representation space.

Extensive experiments show that Slot-ID improves identity preservation and perceptual quality while remaining competitive in prompt following. The gains are particularly apparent in challenging cases involving large facial-expression changes, viewpoint variation, and rapid motion, where single-image conditioning is most prone to identity drift, deformation, or pose locking. These results suggest that short reference videos provide a practically useful conditioning signal for identity-preserving T2V, and that compact temporal identity aggregation is a key ingredient for making such signals effective in modern video generators.

Our main contributions can be summarized as follows.

\begin{itemize}
\item We study identity-preserving text-to-video generation in a video-referential setting, showing that short reference clips provide temporal identity cues unavailable in a single image.
\item We introduce Slot-ID, a slot-based temporal identity encoding framework that distills these cues into a compact and backbone-compatible identity condition.
\item We demonstrate through quantitative and qualitative evaluation that this design improves identity consistency under substantial pose, expression, and motion changes, achieving state-of-the-art (SOTA) identity preservation.
\end{itemize}

\section{Related Work}

\paragraph{Video diffusion backbones and control signals.}
Modern video generation increasingly builds on diffusion transformers (DiT/MMDiT) operating in 3D VAE latents~\cite{esser2024scaling}, largely replacing U-Nets~\cite{ronneberger2015u}. Open backbones such as CogVideoX, HunyuanVideo, and Wan leverage dual-stream attention, 3D VAEs, and large text encoders~\cite{yang2024cogvideox,kong2024hunyuanvideo,wan2025wan}. 
For image generation, Filter-Guided Diffusion~\cite{gu2024filterguided} provides efficient, training-free control over guidance strength and frequencies through fast filtering during denoising.
Rather than attaching heavy control branches or modifying the base~\cite{controlnet++}, \textbf{Slot-ID} derives an identity signal from a short reference clip and injects it into a \emph{frozen} DiT/MMDiT backbone (e.g., Wan~\cite{wan2025wan}) via lightweight conditioning, preserving the base checkpoint and compatibility with existing controllers/schedulers.

\paragraph{Personalization strategies: tuning-based vs.\ tuning-free.}

Identity conditioning generally follows (i) \emph{tuning-based} approaches---per-subject fine-tuning via DreamBooth~\cite{ruiz2023dreambooth}, LoRA adapters~\cite{hu2022lora,biderman2024loralearnsforgets,jin2024conditionalloraparametergeneration}, and textual inversion~\cite{gal2022imageworthwordpersonalizing}---which can achieve high fidelity from few images/clips but add per-identity compute and risk overfitting/background leakage; LoRA/text tokens demand careful hyperparameters to avoid overspecification and motion dampening~\cite{zhong2024multiloracompositionimagegeneration,kumari2023multiconceptcustomizationtexttoimagediffusion,tewel2024keylockedrankeditingtexttoimageperfusion}. 
Numerous ready-to-use identity LoRAs exist in community hubs~\cite{civitai}. (ii) \emph{Tuning-free} methods extract identity embeddings from one/few references---often CLIP/ArcFace or image encoders~\cite{Deng_2022Arcface,radford2021learningtransferablevisualmodelsCLIP,chen2023photoversetuningfreeimagecustomization}---and inject them at inference as cross-attention visual tokens or identity-augmented text tokens~\cite{li2024photomaker,ye2023ip,wang2024instantid,li2023blipdiffusionpretrainedsubjectrepresentation,wei2025echovideoidentitypreservinghumanvideo,ConsisIDPreviewData2024}. 

\paragraph{Identity-preserving video generation.}

Prior work splits into two camps. Per-subject fine-tuning of the backbone or lightweight adapters attains strong fidelity but scales poorly in training/storage~\cite{ruiz2023dreambooth,gal2022imageworthwordpersonalizing,jiang2024videobooth,materzynska2024newmove,wang2025motion}. Tuning-free alternatives infer identity at inference from a reference image using ArcFace/CLIP~\cite{Deng_2022Arcface,radford2021learningtransferablevisualmodelsCLIP} and inject these features into mostly frozen diffusion backbones via cross-attention visual tokens or identity-augmented text tokens~\cite{li2024photomaker,ye2023ip,wang2024instantid,li2023blipdiffusionpretrainedsubjectrepresentation}; open-source variants often add a face adapter to a pretrained T2V model~\cite{he2024idanimatorzeroshotidentitypreservinghuman}, while commercial systems keep models/training proprietary~\cite{polyak2025moviegencastmedia}. Because single-image cues miss fine traits and are brittle under large pose or edits, plug-and-play designs route image features into frozen DiT backbones~\cite{xue2025standinlightweightplugandplayidentity,wei2025echovideoidentitypreservinghumanvideo,yuan2025identitypreservingtexttovideogenerationfrequency}; heavier end-to-end systems (e.g., Phantom) improve consistency at substantially higher cost~\cite{liu2025phantomsubjectconsistentvideogeneration}. We take a video-referential route: a short reference clip is encoded by a temporal identity encoder into identity tokens and injected into a frozen DiT/MMDiT backbone, preserving identity across motion/view/illumination changes, and improving prompt alignment---without per-identity fine-tuning---as in \textbf{Slot-ID}.

\section{Method}

\begin{figure*}[t]
  % \centering
  \includegraphics[width=\textwidth]{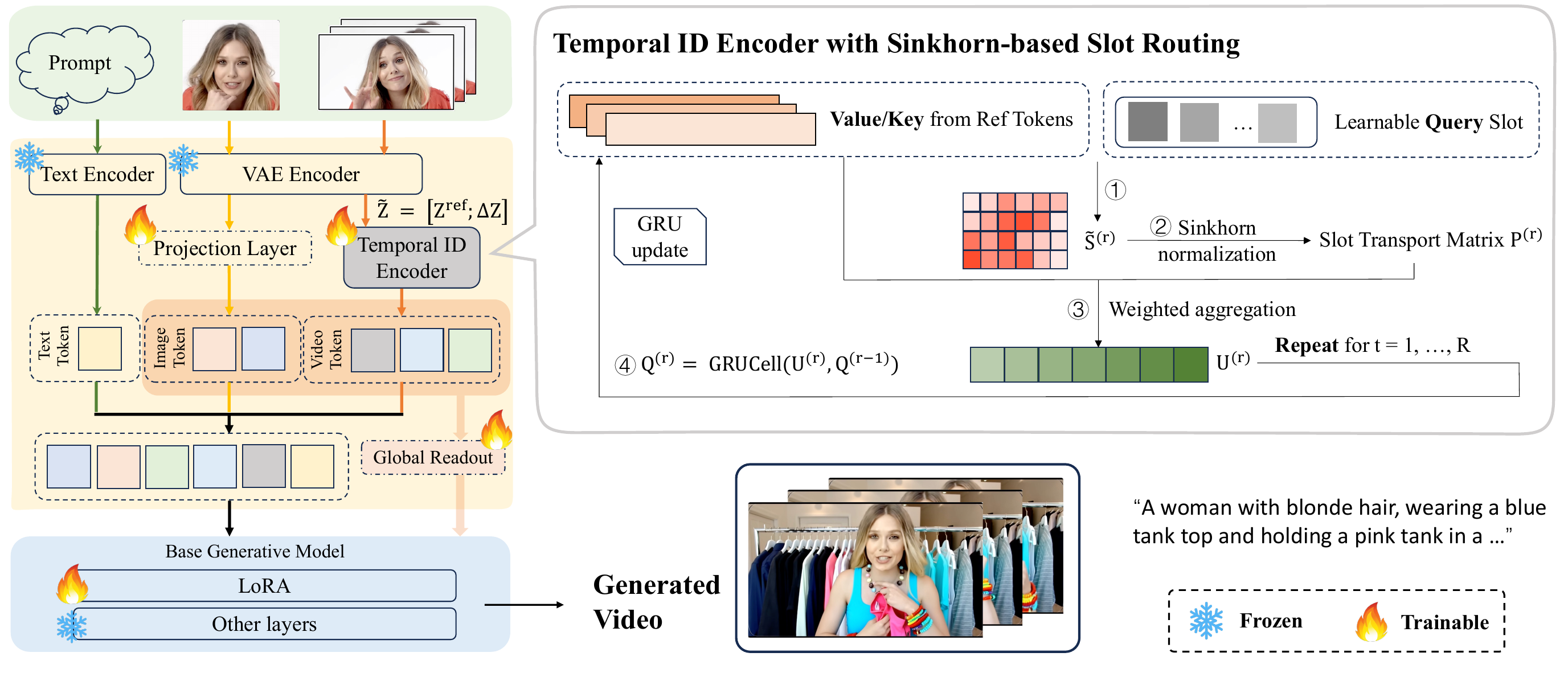} 
  \caption{\textbf{Pipeline overview.} A text prompt, a background-neutral face reference, and a reference video are encoded to provide conditioning signals for generation. A Sinkhorn-routed slot reader then iteratively refines learnable slot queries: (1) compute query--token similarity scores; (2) apply temperature-scaled Sinkhorn normalization to obtain a (near) doubly-stochastic transport matrix; (3) aggregate values; (4) update queries with a GRU to yield identity slots.}
  \Description{The pipeline diagram shows how Slot-ID encodes a text prompt, an image anchor, and a reference video. The reference-video branch uses a temporal encoder and a Sinkhorn-routed slot reader to produce compact identity slots that condition the frozen video generator.}
  \label{fig:maingraph}
\end{figure*}

\subsection{Overview}

We study identity-preserving text-to-video generation conditioned on a short reference clip. Let $\mathcal{V}^{ref}$ denote a reference video of the target person, and let $\mathcal{I}^{ref}$ denote a near-frontal portrait extracted from the same subject. Given a text prompt, $\mathcal{V}^{ref}$, and $\mathcal{I}^{ref}$, our goal is to generate a prompt-faithful video while preserving the target identity across changes in pose, expression, illumination, and motion.

Our framework builds on a frozen Wan text-to-video backbone~\cite{wan2025wan} and introduces a lightweight identity-conditioning module. The role of this module is to convert the reference inputs into a compact identity representation that can be injected into the frozen generator without modifying the backbone architecture. Concretely, the model produces two complementary conditioning streams: a set of temporal identity tokens extracted from the reference clip, and a set of anchor identity tokens extracted from the reference image. These identity tokens are appended to the text conditioning, allowing the backbone to attend to identity information throughout denoising. In parallel, a global identity vector is computed and used for feature-wise modulation. The overall pipeline is shown in Fig. \ref{fig:maingraph}.

\subsection{Slot-ID: Slot-based Temporal Identity Encoder}

We first describe how the reference clip is converted into a compact set of temporal identity tokens. Let $\mathcal{V}^{ref}\in \mathbb{R}^{B\times T\times H_0\times W_0 \times 3}$ be the input reference video, where $B$ is the batch size, $T$ is the number of frames, and $H_0$ and $W_0$ are the frame height and width. Using the same VAE encoder as the frozen video backbone, we map $\mathcal{V}^{ref}$ to latent features $\mathbf{Z}^{ref} \in \mathbb{R}^{B\times C\times T\times H\times W}$, where $C$ is the latent channel dimension and $H$, $W$ are the spatial dimensions in latent space.

To expose short-range temporal variation, we construct a temporal-difference latent volume $\Delta\mathbf{Z}=\{\Delta z_t\}^T_{t=1}$, with
\begin{equation}
\Delta\mathbf{z}_t \ =\
\begin{cases}
\mathbf{0}, & t\ =\ 1,\\[2pt]
\mathbf{z}^{ref}_t - \mathbf{z}^{ref}_{t-1}, & t\ >\ 1,
\end{cases}
\end{equation}
where $\mathbf{z}^{ref}_t \ \in \ \mathbb{R}^{B\times C\times H\times W}$ denotes the latent slice at time $t$. 
We then concatenate the original latent and the temporal-difference latent along the channel dimension, $\tilde{\mathbf{Z}}\ =\ [\mathbf{Z}^{\mathrm{ref}}\!;\,\Delta\mathbf{Z}]$, and feed $\tilde{\mathbf{Z}}$ to the temporal identity encoder. This encoder first applies a strided 3D patch embedding with temporal and spatial kernel sizes $(k_t, h, w)$, producing a token sequence $\mathbf{X} \ \in \ \mathbb{R}^{B\times L\times D}$, where $D$ is the token dimension and 
$L \ =\ \frac{T}{k_t}\cdot\frac{H}{h}\cdot\frac{W}{w}$
is the number of spatiotemporal tokens. 
A small stack of spatiotemporal self-attention (STSA) blocks further mixes these tokens before identity readout using 3D RoPE and Flash-style kernels \cite{dao2022flashattentionfastmemoryefficientexact,su2023roformerenhancedtransformerrotary}.

The remaining question is how to summarize the token sequence $X$ into a small number of identity tokens. Rather than pooling all clip tokens into a single vector, we represent the clip with $S$ learnable slots, where $S$ is the number of output identity tokens. Each slot acts as a query that reads a compact identity summary from the spatiotemporal token sequence. This yields a compact temporal identity code $\mathbf{C}_{\mathrm{id}}\ \in  \ \mathbb{R}^{B\times S\times D}$, which is later used to condition the generator.

\subsubsection{Sinkhorn-Routed Slot Reader}
\label{sec:sinkhorn_reader}

Let $\mathbf{X} \ \in \ \mathbb{R}^{B\times L\times D}$ be the output token sequence of the temporal encoder. We use the same tensor as keys and values, $K=X$, $V=X$, where $K,V\in \mathbb{R}^{B\times L\times D}$. The slot reader maintains $S$ learnable slot states $Q^{(r)} \ \in \ \mathbb{R}^{B\times S\times D}$, where $r\in\{0,1,...,R\}$ indexes the refinement step and $R$ is the total number of refinement iterations. The initial slot states $Q^{(0)}$ are learned parameters shared across samples.

At iteration $r$, we compute slot-token similarity scores
\begin{equation}
A^{(r)}\ =\ \frac{Q^{(r-1)}(K)^\top}{\sqrt{D}}\ \in\ \mathbb{R}^{B\times S\times L}.
\end{equation}
To control the sharpness of the assignment during training, we divide the scores by a temperature $\tau(s)$, where $s$ denotes the global training step:
$\tilde{A}^{(r)}\ =\ \frac{A^{(r)}}{\tau(s)}$.
The temperature follows a linear schedule
$\tau(s)\ =\ \tau_{\text{start}}+\min\!\Big(1,\frac{s}{T_{\text{decay}}}\Big)\cdot\big(\tau_{\text{end}}-\tau_{\text{start}}\big)$,
where $\tau_{start}$ and $\tau_{end}$ are the initial and final temperatures, and $T_{decay}$ is the number of steps used for annealing.

Rather than using a standard softmax assignment, we compute the slot-token correspondence with an entropic optimal-transport projection under uniform marginals, which encourages a more balanced allocation of reference evidence across slots. Combined with temperature annealing, this yields broader assignments in the early stage of training and progressively sharper correspondences later on. Formally, the assignment matrix
\begin{equation}
P^{(r)} = \operatorname{Sinkhorn}\!\left(\tilde{A}^{(r)}\right)
\in \mathbb{R}^{B\times S\times L},
\end{equation}
is obtained by applying Sinkhorn normalization~\cite{cuturi2013sinkhorndistanceslightspeedcomputation,ConcerningNon,peyr2020computationaloptimaltransport} to the scaled similarity matrix \(\tilde{A}^{(r)}\) in the log domain, where \(P^{(r)}\) denotes the approximately doubly-stochastic coupling between the \(S\) slots and the \(L\) reference tokens.

Specifically, we set
$\log K_{\mathrm{sh}} = \tilde{A}^{(r)}$,
and use uniform target marginals
$a = \frac{1}{S}\mathbf{1}_{S}$, $b = \frac{1}{L}\mathbf{1}_{L}$,
where \(\mathbf{1}_{S}\) and \(\mathbf{1}_{L}\) are all-ones vectors of length \(S\) and \(L\), respectively. The dual variables are initialized as
$\log u = \mathbf{0}$, $\log v = \mathbf{0}$,
and updated for \(t=1,\dots,n_{\mathrm{iter}}\) as
\begin{equation}
\begin{aligned}
\log u &\leftarrow \log a - \operatorname{logsumexp}\!\big(\log K_{\mathrm{sh}} + \log v\big)_{\mathrm{col}},\\
\log v &\leftarrow \log b - \operatorname{logsumexp}\!\big(\log K_{\mathrm{sh}} + \log u\big)_{\mathrm{row}}.
\end{aligned}
\end{equation}
After the final iteration, the log-coupling is assembled as
\begin{equation}
\log P^{(r)} \ =\ \log K_{\text{sh}} + \log u + \log v,
\end{equation}
and the assignment matrix is obtained by exponentiation:
$P^{(r)}=\exp(\log P^{(r)})$.
Because Sinkhorn normalization is performed for a finite number of floating-point iterations, the resulting coupling may deviate slightly from exact doubly-stochastic constraints. We therefore apply a final row-and-column renormalization,
\begin{equation}
P_{i,:}^{(r)} \leftarrow \frac{P_{i,:}^{(r)}}{\sum_j P_{ij}^{(r)}+\varepsilon},\quad
P_{:,j}^{(r)} \leftarrow \frac{P_{:,j}^{(r)}}{\sum_i P_{ij}^{(r)}+\varepsilon},
\end{equation}
where $\varepsilon$ is a small constant for numerical stability. This yields an approximately doubly-stochastic assignment matrix, which is then used to aggregate slot features.

Given the assignment matrix, we compute slot-wise aggregated features
$U^{(r)} \ =\  P^{(r)} V \ \in\ \mathbb{R}^{B\times S\times D}$.
These aggregated features are used to update the slot states with a recurrent cell:
$Q^{(r)} \ =\  \mathrm{GRUCell}(U^{(r)},\,Q^{(r-1)})$. After $R$ refinement steps, the final slot states are projected to produce the temporal identity tokens:
$\mathbf{C}_{\mathrm{id}}\ =\ \mathrm{Proj}(Q^{(R)})\ \in\ \mathbb{R}^{B\times S\times D}$, where $\mathrm{Proj}(\cdot)$ denotes a LayerNorm-Linear readout head. The resulting $\mathbf{C}_{\mathrm{id}}$ serves as the video-conditioned identity representation used by the generator.

\subsection{Dual-source identity conditioning}

The temporal tokens $\mathbf{C}_{\mathrm{id}}$, which are extracted from the reference clip and capture motion-stable, person-specific cues, are complemented by an image-based identity stream. Let $I_{ref}$ be a background-neutral portrait of the same person. After projection to the backbone token space, this image branch produces  $\mathbf{C}_{\mathrm{img}} \ \in \  \mathbb{R}^{B\times K\times D}$ as \emph{image-anchor} tokens, where $K$ is the number of image identity tokens and $D$ matches the backbone conditioning dimension.

We combine the two identity streams with a scalar gate $w  \in   [0,1]$ during both training and inference:
\begin{equation}
\widehat{\mathbf{C}}_{\mathrm{id}}\ =\ (1-w)\,\mathbf{C}_{\mathrm{id}},\qquad
\widehat{\mathbf{C}}_{\mathrm{img}}\ =\ w\,\mathbf{C}_{\mathrm{img}}.
\end{equation}
The gate $w$ follows a schedule during denoising. At earlier denoising stages, a larger $w$ places more emphasis on the clean, scene-agnostic identity anchors from the reference image; at later stages, a smaller $w$ gradually shifts the conditioning toward the temporal identity cues extracted from the reference clip. In this way, the two streams play complementary roles over the course of generation rather than contributing with a fixed static ratio.
The final identity prefix is formed by concatenating the weighted image tokens and weighted video tokens, and then prepending them to the text tokens:
\begin{equation}
\mathbf{C}_{prefix}=[\widehat{\mathbf{C}}_{\mathrm{img}};\widehat{\mathbf{C}}_{\mathrm{id}};\mathbf{C}_{\mathrm{text}}],
\end{equation}
where $\mathbf{C}_{\mathrm{text}}$ denotes the text-token sequence produced by the text encoder. 

In parallel, we compute one global vector from each branch, $\mathbf{g}_{\mathrm{vid}}\in\mathbb{R}^{B\times D}$, $\mathbf{g}_{\mathrm{img}}\in\mathbb{R}^{B\times D}$,
and combine them with the same gate:
$\mathbf{g} \ =\ (1-w)\,\mathbf{g}_{\mathrm{vid}} \;+\; w\,\mathbf{g}_{\mathrm{img}}$.
The fused vector $g$ is then used for FiLM-style modulation in the backbone, providing a continuous global identity prior in addition to the token prefix.

\subsection{Training}

The Wan backbone remains frozen during training. Identity control is introduced only through the added conditioning modules and lightweight trainable adapters. At each denoising step, the text tokens $\mathbf{C}_{\mathrm{text}}$, image identity tokens $\mathbf{C}_{\mathrm{img}}$, and temporal identity tokens $\mathbf{C}_{\mathrm{id}}$ are combined into the conditioning sequence described above. To reduce over-conditioning, we apply prefix-token dropout and lightweight regularization before cross-attention.
A global identity vector $g$ provides gentle FiLM-style modulation to temporal features. For stability, we insert LoRA on cross-attention $K/V$ projections and keep all base weights frozen.

Training follows the base model's latent-space velocity/flow-matching ($v$-prediction) objective \cite{wan2025wan}: for VAE latents $z$, sample $\varepsilon\!\sim\!\mathcal{N}(0,I)$ and $t\!\sim\!\mathcal{U}[0,1]$, set $z_1 \ = \ z$, $z_0 \ = \ \varepsilon$, $z_t \ = \ (1-t)z_1+t z_0$, and target $v^*(z_t) \ = \ z_0-z_1$. We minimize:
\begin{equation}
\mathcal{L} \ = \ \mathbb{E}_{z,\varepsilon,t}\big\|v_{\theta}(z_t,t,\mathbf{C}_{prefix},g)-(z_0-z_1)\big\|_2^2.
\end{equation}

\section{Results and Evaluation}

\begin{figure*}[t]
  \centering
  \includegraphics[width=\textwidth]{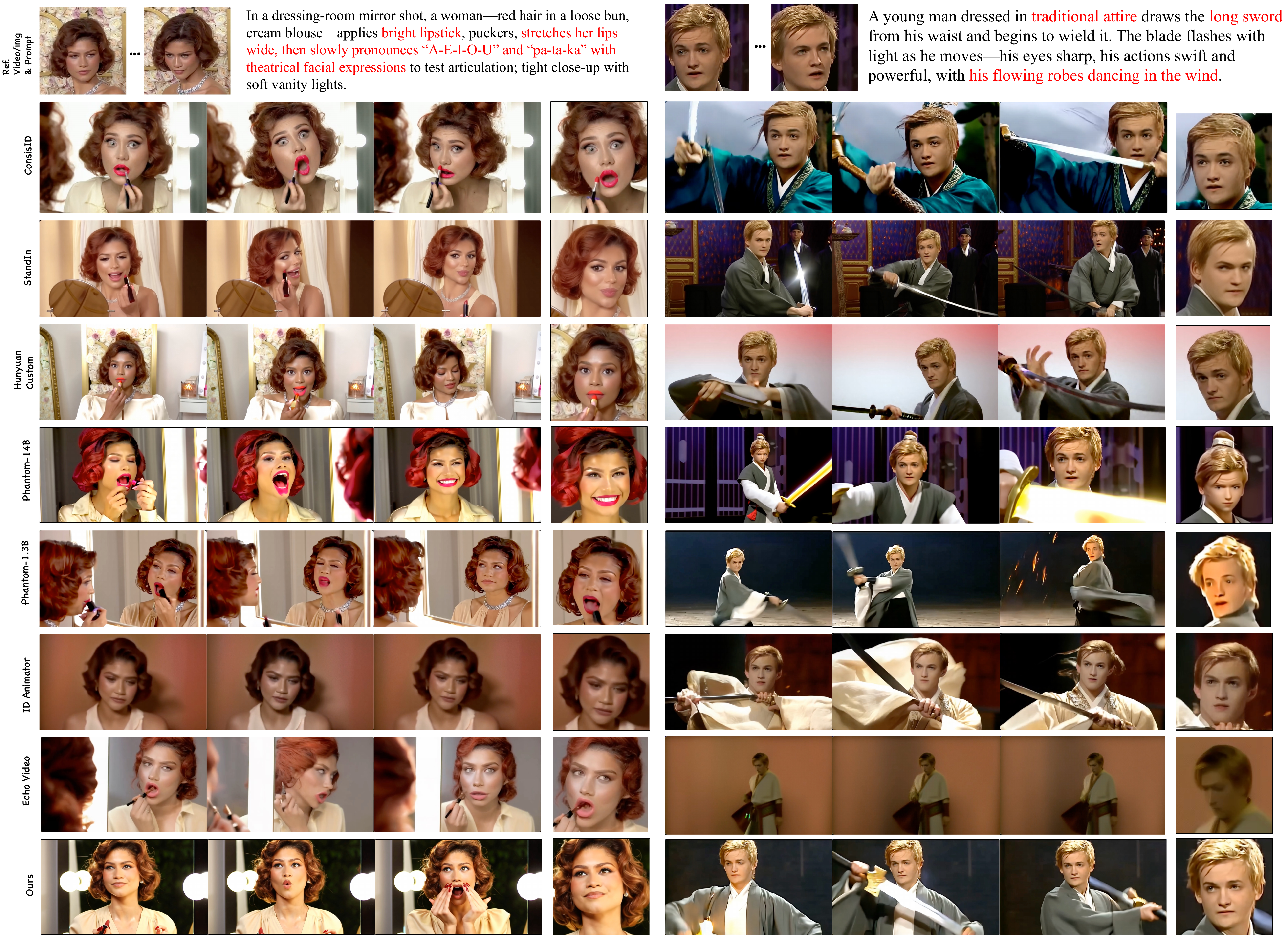}
  \caption{\textbf{Identity-preserving video generation.}
   Top: reference video (3 frames; leftmost used by image-conditioned baselines) and prompt; rows: each method with face crops. Baselines exhibit identity drift, expression/lip misalignment, over-smoothing, and flicker under non-frontal poses and background motion; ours preserves facial structure and expression/lip coherence, yielding sharper, temporally stable frames with consistent backgrounds. Best viewed zoomed in.}
  \Description{The qualitative comparison shows reference frames and prompts at the top, followed by generated frames and face crops from different methods. Slot-ID maintains more consistent facial identity and temporal appearance than the compared baselines across non-frontal poses and background motion.}
  \label{fig:qualitative}
\end{figure*}

\subsection{Experimental Setup}

\paragraph{Training Setup.}
We train our model on the publicly released human-centric dataset~\cite{ConsisIDPreviewData2024} introduced in \cite{yuan2025identitypreservingtexttovideogenerationfrequency} used in prior identity-preserving video generation work.
For each training sample, the target is a full 65-frame clip at 16 fps. The reference video is a clear, temporally stable face track of the same identity, selected from a different temporal interval of the same source video whenever possible. Its length is adaptively set to 17--33 frames, roughly 1--2 seconds, and we shorten the reference rather than include blurry or unstable frames when a high-quality 33-frame track is unavailable. The reference track is obtained by detecting the dominant subject, consolidating detections into a stable square face crop, and resizing all frames. This reference crop provides local facial identity evidence, while the target remains the captioned full video, so training learns identity transfer rather than reconstruction of the reference segment.
In parallel, we construct a reference identity image by sampling candidate frames from the reference track, cropping the dominant face, parsing a background-free portrait at 512$\times$512, and selecting the best candidate according to a joint quality score that favors sharpness, near-frontal view, and neutral expression. We fine-tune a Wan-2.1 T2V-14B~\cite{wan2025wan} backbone with all base parameters frozen, adapting only the cross-attention layers with LoRA on the key/value projections. The temporal identity encoder uses 6 slots, 3 refinement iterations, 2 STSA layers, 16 heads, per-token dropout of 0.05, 2 image-anchor tokens  and a global branch for FiLM modulation on the last 50\% of backbone blocks. Additional implementation details, reference-length sensitivity, and degraded-reference robustness studies are provided in the supplementary material.

\paragraph{Evaluation benchmark.} 
Our evaluation is restricted to human identity-preserving T2V. The evaluation set is constructed separately from additional CC-licensed YouTube videos collected outside the training set, with held-out identities and source videos. For each evaluation instance, we sample a short reference video, extract its sharpest frontal frame as the image cue for image-conditioned baselines, and pair it with a prompt designed to require substantial appearance-preserving motion or viewpoint change.

\paragraph{Metrics.} Following Stand-In \cite{xue2025standinlightweightplugandplayidentity}, we report three complementary metrics, all higher-is-better. \emph{Face Similarity} measures identity preservation as the embedding similarity between the reference face and detected faces in generated frames, averaged over frames. \emph{Naturalness} uses GPT\textendash4o to rate perceptual realism from uniformly sampled frames under a fixed $1$--$5$ rubric with OpenS2V protocol \cite{yuan2025opens2v}; we report the mean score. \emph{Prompt Following} evaluates text--video alignment by encoding the prompt and a clip with XCLIP (base--patch32)~\cite{ma2022x} and taking the similarity.

\paragraph{Baselines.} We compare \textbf{Slot-ID} against representative open-source identity-preserving video generation methods, including ConsisID, Stand-In, HunyuanCustom, Phantom, ID-Animator, and EchoVideo. For each case, we sample a reference video, take its sharpest frontal frame as the image cue, and draw the prompt at random. All baselines are run with public checkpoints under a unified evaluation protocol.

\subsection{Comparison}

\paragraph{Quantitative Comparison}
Table~\ref{table1} summarizes the quantitative comparison. Slot-ID achieves the best Face Similarity and Naturalness scores among all compared methods, while remaining close to the best Prompt Following score. These results indicate that the gains in identity preservation are not achieved by sacrificing prompt fidelity. Instead, our model improves identity consistency while maintaining strong alignment with the input instruction.
The overhead remains modest: because the frozen Wan backbone dominates inference and the reference encoder runs only once per identity, Slot-ID adds 5.1\% end-to-end inference time and about 2GB peak GPU memory over the frozen backbone.

\paragraph{Qualitative Comparison}

\begin{figure}[t]
  \centering
  \includegraphics[width=\linewidth]{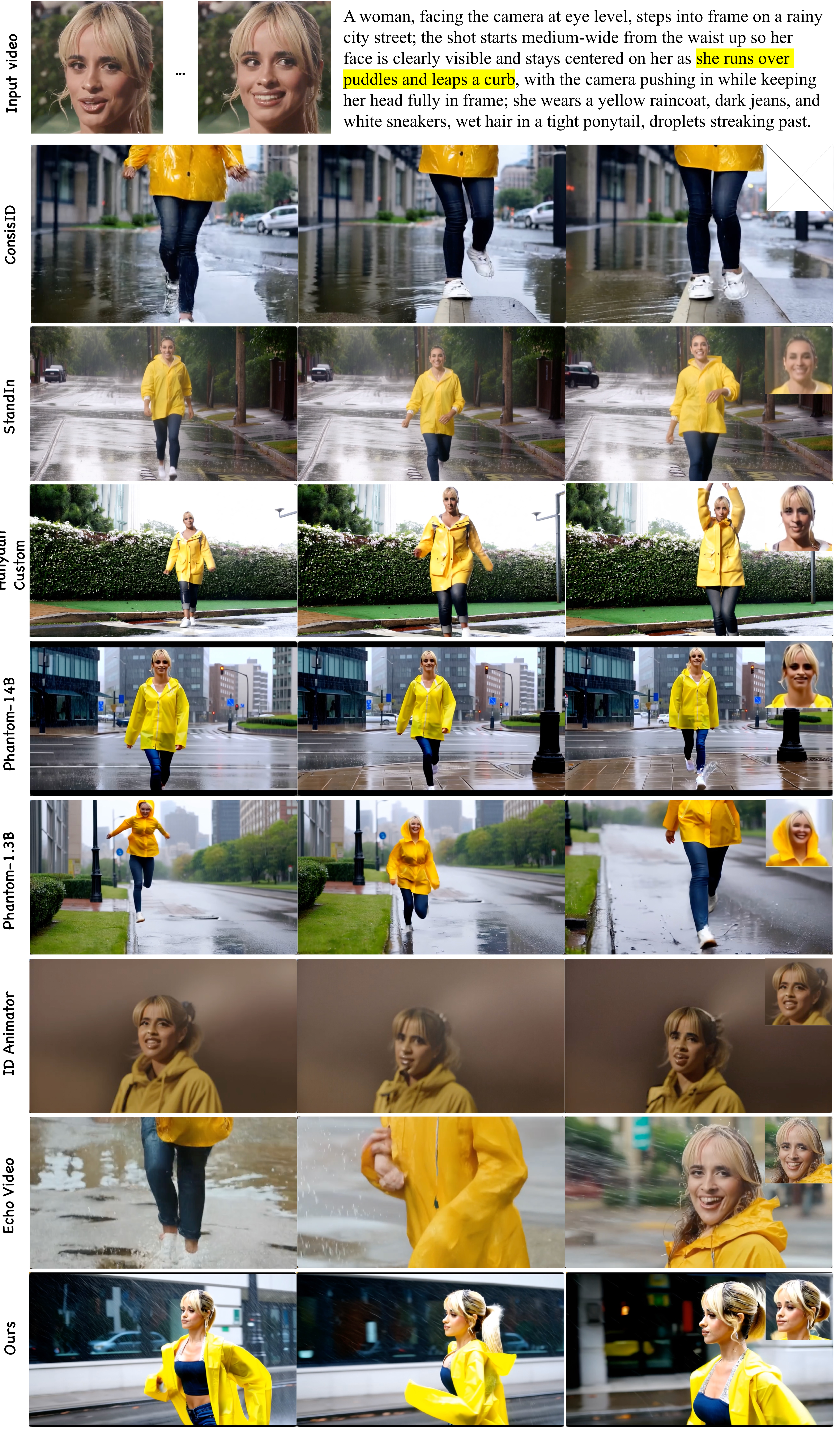}
  \caption{\textbf{Large-motion stress test.} Baselines often hide the face or show identity drift/deformation; \textbf{ours} performs well.}
  \Description{The large-motion stress test compares generated results under prompts with substantial body or camera motion. Baseline methods often hide or distort the face, while Slot-ID better preserves visible facial identity.}
  \label{fig:bigmove}
\end{figure}

Qualitative results are shown in Figs.~\ref{fig:qualitative} and \ref{fig:bigmove}. Across diverse prompts, Slot-ID more reliably preserves facial structure, local identity traits, and temporal consistency under substantial changes in viewpoint, motion, and expression. In contrast, the baselines exhibit several recurring failure modes. ConsisID tends to ignore the target identity and drift toward a prompt-biased generic face. ID-Animator often fails in full-body or complex-scene settings. Stand-In and EchoVideo can preserve coarse appearance but frequently lose fine identity cues and sometimes inherit clothing or pose details from the reference. HunyuanCustom shows identity washout under strong actions or camera motion. Phantom-14B, while strongest among the baselines, still becomes prone to facial warping or off-model identity under rapid motion and large view changes.

These differences become especially pronounced in \textbf{stress cases}. Following DrawBench~\cite{saharia2022photorealistictexttoimagediffusionmodelsDrawBench} and recent T2V benchmarks~\cite{huang2023vbenchcomprehensivebenchmarksuite,sun2025t2vcompbenchcomprehensivebenchmarkcompositional}, we group test cases into four challenge factors: (1) large whole-body or viewpoint/camera motion, (2) exaggerated facial expressions, (3) multi-object interactions, and (4) partial occlusions. Across all four, Slot-ID more consistently preserves identity and temporal coherence, while baselines more often show identity drift, facial deformation, or background inconsistency (Fig.~\ref{fig:bigmove}; additional results in supplementary). This suggests Slot-ID's advantage is not merely a metric gain, but a more robust identity preservation precisely where temporal cues matter most.

\begin{table}[t]
  \centering
  \small
  \setlength{\tabcolsep}{3pt}
  \renewcommand{\arraystretch}{1.05}
  \begin{tabular*}{\columnwidth}{@{\extracolsep{\fill}}lccc}
    \toprule
    Method & Face Sim.\,$\uparrow$ & Naturalness\,$\uparrow$ & Prompt Following \\
    \midrule
    ID-Animator      & 0.298 & 2.897 & 0.560 \\
    HunyuanCustom   & 0.643 & 3.423 & 0.604 \\
    ConsisID       & 0.418 & 3.107 & \underline{0.640} \\
    Stand-In         & 0.697 & 3.887 & 0.611 \\
    EchoVideo       & 0.671 & 3.856 & 0.635 \\
    Phantom-1.3B    & 0.467 & 3.478 & \textbf{0.643} \\
    Phantom-14B     &\underline{0.699} & \underline{3.912} & 0.615 \\
    \addlinespace[2pt]
    \textbf{Slot-ID (Ours)} & \textbf{0.729} & \textbf{3.917} & 0.634 \\
    \bottomrule
  \end{tabular*}
  \caption{Quantitative comparison with identity-preserving video generation methods. Higher is better; best and second-best are in \textbf{bold} and \underline{underlined}.}
  \label{table1}
\end{table}

\paragraph{Pose locking.}
A specific failure mode of image-conditioned methods is \emph{pose locking}~\cite{xue2025standinlightweightplugandplayidentity}: the subject's pose in the reference image exerts an overly strong prior on generation, causing the synthesized subject to remain anchored to the reference posture while still attempting to follow the prompt---producing a forced, physically implausible compromise where the action is partially executed but in a contorted, unnatural way no real person would adopt. As illustrated in the supplementary material, Stand-In~\cite{xue2025standinlightweightplugandplayidentity} exhibits this artifact clearly: in Prompt~1 (playing piano), the subject faces the camera throughout rather than turning toward the keys; in Prompt~2 (tying shoelaces), the head remains upright and frontal instead of tilting downward---yielding physically implausible poses. Slot-ID instead generates poses that are natural and consistent with the prompted action, free from reference-posture anchoring. More pronounced pose-locking examples across a wider range of baselines are provided in the supplementary video.

\subsection{User Study}
To complement numerical evaluation, we conducted a human evaluation on identity-conditioned T2V generation. We sampled $15$ identities from the benchmark and generated one clip per method under matched prompts. More than $50$ participants rated anonymized, order-randomized comparisons on three $1$--$5$ Likert scales: \emph{Face Similarity}, \emph{Visual Quality}, and \emph{Text Alignment}. Mean Opinion Scores (MOS) were averaged across raters and identities.
As summarized in Table~\ref{tab:userstudy}, our method achieves the highest MOS on all three criteria, with particularly strong gains on \emph{Face Similarity} and \emph{Visual Quality} under challenging motion and occlusion prompts, consistent with the intended role of the proposed conditioning strategy. Notably, Slot-ID also achieves the highest human-rated Text Alignment, aligning with the quantitative finding that the method remains competitive on prompt following despite prioritizing identity preservation. In practice, this suggests that the additional temporal identity signal improves perceptual coherence without making the generator less responsive to the text prompt.

\begin{table}[t]
\centering
\footnotesize
\setlength{\tabcolsep}{4pt}
\begin{tabular*}{\columnwidth}{@{\extracolsep{\fill}}lccc}
\toprule
Method & Face Similarity & Visual Quality & Text Alignment \\
\midrule
ConsisID        & 2.515 & 2.535 & 2.842 \\
EchoVideo       & 2.856 & 2.614 & 2.769 \\
ID-Animator      & 2.447 & 1.727 & 1.644 \\
Phantom-14B    & 3.462 & 3.345 & 3.258 \\
Phantom-1.3B   & 3.352 & 3.061 & 3.098 \\
Stand-In         & 3.199 & 3.249 & 3.131 \\
HunyuanCustom   & 3.197 & 2.871 & 3.008 \\
Ours            & \textbf{3.837} & \textbf{3.890} & \textbf{3.639} \\
\bottomrule
\end{tabular*}
\caption{\textbf{User study.} Over $50$ raters evaluated videos for $15$ identities on three 5-point criteria---\emph{Face Similarity}, \emph{Visual Quality}, \emph{Text Alignment}; numbers are mean opinion scores (higher is better). Ours ranks first on all.}
\label{tab:userstudy}
\end{table}

\subsection{Ablation Study}

\paragraph{Effectiveness of Slot-based Identity Encoding.}

We conduct ablation studies to examine which parts of the proposed identity-conditioning design matter in practice. Unless otherwise stated, all ablation variants are retrained from scratch under the same training recipe as the main model, with matched training data, optimization schedule, LoRA policy, and injection interface.

\textbf{Effect of video-referential identity evidence.}
We first isolate the role of the reference-video signal. We compare Slot-ID with two controlled variants: a repeated-image pseudo-video that tiles the reference portrait across all video frames, and an image-anchor-only variant that removes the video-token pathway while keeping the same frozen backbone and image identity stream. As shown in Table~\ref{tab:additional_ablations}, the real reference video yields the best Face Similarity and Naturalness. This gap confirms that real temporal and multi-view reference evidence provides identity cues unavailable from a static portrait.

\begin{figure}[t]
  \centering
  \includegraphics[width=\linewidth]{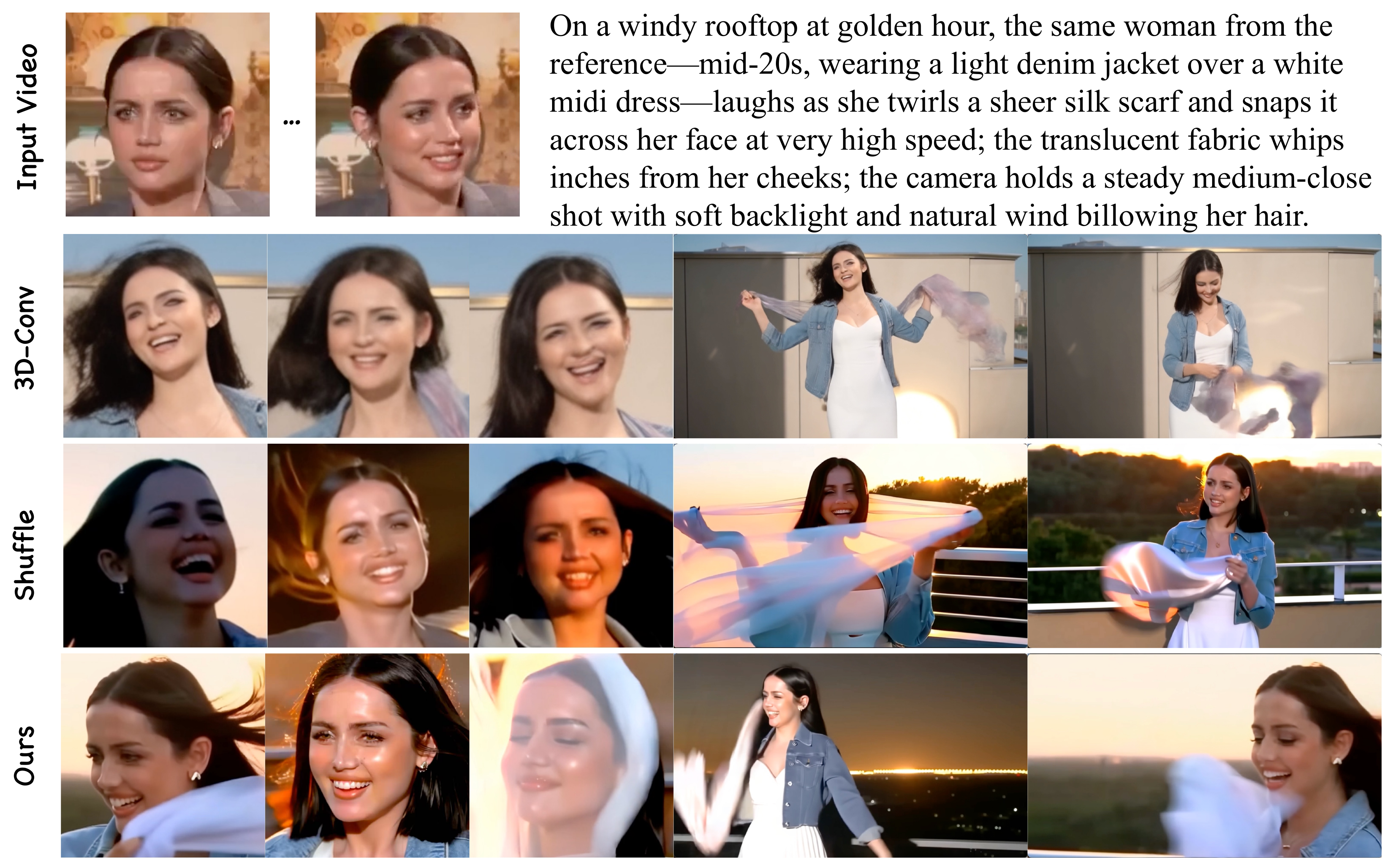}
  \caption{\textbf{Ablation study.} Two variants drift and flicker, while ours maintains identity and stable contours across occlusions and large pose changes.}
  \Description{The ablation figure compares visual results from two ablated variants and the full Slot-ID model. The variants show identity drift and flickering contours, while the full model maintains more stable identity across occlusions and large pose changes.}
  \label{fig:ablation}
\end{figure}

\textbf{Naive video encoding vs. slot-based identity encoding.} We then compare our temporal identity encoder with a 3DConv-Pool variant that replaces the slot-based reader with a 3D convolutional encoder followed by global pooling to the same number of output tokens. As shown quantitatively in Table~\ref{tab:additional_ablations} and qualitatively in Fig.~\ref{fig:ablation}, this variant weakens identity preservation and produces more look-alikes, blur, and shape/texture artifacts, especially under camera motion and pose change. This supports the claim that naive compact video encoding tends to average away fine person-specific cues that are better preserved by slot-based aggregation.

\textbf{Effect of temporal order.} We test whether ordered temporal evidence is necessary by shuffling the sampled reference frames before feeding them into the temporal identity encoder. This orderless-reference variant consistently degrades identity robustness and increases temporal jitter. The effect is particularly visible in cases with occlusion and rapid head motion, as shown in Fig.~\ref{fig:ablation}, indicating that the model benefits from structured temporal evidence rather than from a mere unordered set of frames.

\textbf{Effect of the temporal-difference branch.} To isolate the contribution of the temporal-difference branch, we remove it while keeping all other training settings identical. The resulting model drops from 0.729 to 0.711 on Face Similarity and from 3.917 to 3.910 on Naturalness, as shown in Table~\ref{tab:additional_ablations}. This suggests that explicit motion-difference cues are helpful for identity preservation, although their contribution is more modest than that of the overall slot-based reader.

\textbf{Effect of Sinkhorn routing.} We replace the Sinkhorn-based reader with a standard softmax reader while keeping the rest of the architecture unchanged. Performance drops substantially: Face Similarity decreases from 0.729 to 0.517, Naturalness from 3.917 to 3.225, and Prompt Following from 0.634 to 0.616, as shown in Table~\ref{tab:additional_ablations}. This result is consistent with our intuition: in our setting, softmax assignment tends to over-concentrate a few strong responses and yields unstable slot-token associations across highly similar neighboring frames, which in turn harms identity stability.

\textbf{Effect of iterative refinement.} We evaluate whether the reader must be iterative, or whether a one-shot Sinkhorn pooling step is sufficient. Removing iterative refinement while keeping Sinkhorn assignment and slot initialization unchanged leads to a pronounced degradation: Face Similarity falls from 0.729 to 0.405, Naturalness from 3.917 to 2.934, and Prompt Following from 0.634 to 0.596, as shown in Table~\ref{tab:additional_ablations}. This confirms that the proposed reader should be understood as an iterative assignment--aggregation--refinement process rather than as a single Sinkhorn layer.

\begin{table}[t]
\centering
\footnotesize
\setlength{\tabcolsep}{2.5pt}
\begin{tabular}{llccc}
\toprule
Category & Method & Face Sim.\,$\uparrow$ & Natural.\,$\uparrow$ & Prompt Fol.\,$\uparrow$ \\
\midrule
Reference signal & Image-anchor only & 0.642 & 3.702 & 0.639 \\
Reference signal & Repeated-image video & 0.658 & 3.812 & 0.637 \\
Reference signal & Ref. video (Ours) & 0.729 & 3.917 & 0.634 \\
\midrule
Temporal encoder & 3DConv-Pool & 0.622 & 3.738 & 0.628 \\
Temporal encoder & Slot reader (Ours) & 0.729 & 3.917 & 0.634 \\
\midrule
Temporal cue design & Full Slot-ID & 0.729 & 3.917 & 0.634 \\
Temporal cue design & w/o Temp.\ Diff. & 0.711 & 3.910 & 0.638 \\
\midrule
Assignment norm. & Sinkhorn (Ours) & 0.729 & 3.917 & 0.634 \\
Assignment norm. & Softmax Reader & 0.517 & 3.225 & 0.616 \\
\midrule
Reader dynamics & Full Iterative (Ours) & 0.729 & 3.917 & 0.634 \\
Reader dynamics & One-Shot Sinkhorn & 0.405 & 2.934 & 0.596 \\
\bottomrule
\end{tabular}
\caption{Ablation results. Real reference videos outperform image-only and repeated-image controls; the slot reader outperforms naive 3DConv pooling; temporal-difference cues bring modest gains, while Sinkhorn routing and iterative refinement are critical.}
\label{tab:additional_ablations}
\end{table}

\section{Conclusion}

We study identity-preserving text-to-video generation from short reference videos. Compared with single-image conditioning, a short reference clip provides temporal identity cues that are important for maintaining identity under large pose and expression changes. To exploit these cues under a limited token budget, we propose Slot-ID, which aggregates video identity evidence into a compact conditioning code for a frozen video generator. Experiments show that Slot-ID improves identity preservation and visual quality, especially in challenging cases with substantial motion and viewpoint variation. Current limitations include the single-subject human/facial identity scope, reliance on relatively clean reference clips and reduced robustness under severe occlusion or very long-horizon generation.

\begin{acks}
The authors would like to thank the reviewers for their insightful comments. This work is supported by NSF China (62322209 \& 62421003), and the XPLORER PRIZE.
\end{acks}

\bibliographystyle{ACM-Reference-Format}
\bibliography{sample-base}

% \appendix









\end{document}